\documentclass[conference]{IEEEtran}
\IEEEoverridecommandlockouts
\usepackage{cite}
\usepackage{amsmath,amssymb,amsfonts}
\usepackage{algorithm}
\usepackage{algorithmic}
\usepackage{subcaption}

\usepackage{graphicx} 
\usepackage{booktabs} 
\usepackage{textcomp}
\usepackage{xcolor}
\usepackage{hyperref}
\usepackage{xcolor,tcolorbox}
\usepackage{amsmath}

\def\BibTeX{{\rm B\kern-.05em{\sc i\kern-.025em b}\kern-.08em
    T\kern-.1667em\lower.7ex\hbox{E}\kern-.125emX}}
\begin{document}

\title{Evaluating LLM Reasoning for Suicide Screening with the Columbia-Suicide Severity Rating Scale}

\author{\IEEEauthorblockN{1\textsuperscript{st} Avinash Patil}
\IEEEauthorblockA{\textit{Ira A. Fulton Schools of Engineering} \\
\textit{Arizona State University}\\
Tempe, USA \\
apatil20@asu.edu}
\and
\IEEEauthorblockN{2\textsuperscript{nd} Siru Tao}
\IEEEauthorblockA{\textit{Heinz College of Information Systems and Public Policy} \\
\textit{Carnegie Mellon University}\\
Pittsburgh, USA \\
sirutao@andrew.cmu.edu}
\and
\IEEEauthorblockN{3\textsuperscript{rd} Amardeep Kour Gedhu}
\IEEEauthorblockA{\textit{School of Education and Counseling Psychology} \\
\textit{Santa Clara University}\\
Santa Clara, USA \\
agedhu@scu.edu}
}

\maketitle

\begin{abstract}
Suicide prevention remains a critical public health challenge. While online platforms such as Reddit’s r/SuicideWatch have historically provided spaces for individuals to express suicidal thoughts and seek community support, the advent of large language models (LLMs) introduces a new paradigm—where individuals may begin disclosing ideation to AI systems instead of humans. This study evaluates the capability of LLMs to perform automated suicide risk assessment using the Columbia-Suicide Severity Rating Scale (C-SSRS). We assess the zero-shot performance of six models—including Claude, GPT, Mistral, and LLaMA—in classifying posts across a 7-point severity scale (Levels 0–6). Results indicate that Claude and GPT closely align with human annotations, while Mistral achieves the lowest ordinal prediction error. Most models exhibit ordinal sensitivity, with misclassifications typically occurring between adjacent severity levels. We further analyze confusion patterns, misclassification sources, and ethical considerations, underscoring the importance of human oversight, transparency, and cautious deployment. Full code and supplementary materials are available at \url{https://github.com/av9ash/llm_cssrs_code}.
\end{abstract}

\begin{IEEEkeywords}
Suicide risk assessment, Large Language Models, C-SSRS, mental health, zero-shot learning, natural language processing, ethical AI, Reddit, digital triage, ordinal classification.
\end{IEEEkeywords}

\section{Introduction}
Suicide is a leading cause of death worldwide, with over 700,000 deaths occurring each year according to the World Health Organization (WHO)\cite{who2021suicide}. Early identification and intervention for individuals at risk of suicide are crucial for effective prevention. Social media platforms have emerged as significant outlets where individuals express suicidal thoughts and seek peer support. Among these, Reddit's r/SuicideWatch serves as an anonymous community where users share their struggles, providing a valuable resource for analyzing suicidal ideation using computational methods.

Traditional suicide risk assessment methods rely on clinical evaluations and structured psychological interviews \cite{posner2011columbia}. However, given the scale and anonymity of social media data, manual screening and intervention remain infeasible. This has driven research interest in automated methods for suicide risk assessment using Natural Language Processing (NLP) and machine learning techniques \cite{coppersmith2018natural, burnap2015machine}. In particular, Large Language Models (LLMs) have demonstrated significant capabilities in understanding and classifying human emotions, making them suitable for evaluating suicide risk from textual data.

The Columbia-Suicide Severity Rating Scale (C-SSRS) is a widely used and validated framework for assessing suicidal ideation and behavior \cite{columbia_cssrs}. While prior work has explored suicide risk classification in mental health research \cite{gaur2019knowledge}, limited studies have applied structured clinical assessment tools like C-SSRS in conjunction with LLMs. This paper aims to bridge this gap by leveraging LLMs for evaluating r/SuicideWatch posts based on the C-SSRS scale.

The key contributions of this work are:
\begin{itemize}
\item The application of C-SSRS for structured suicide risk assessment in online social media posts.
\item The use of zero-shot learning with Large Language Models for automated classification of suicide risk levels.
\item Comparison of Model performance with traditional NLP techniques and human annotations.
\item Limitations of using LLMs for suicide risk screening.
\end{itemize}

The remainder of this paper is structured as follows: Section 2 discusses related work in suicide risk assessment using NLP. Section 3 presents the methodology, including data preprocessing and model implementation. Section 4 details experimental results and evaluation metrics. Section 5 provides conclusions and directions for future work.

\section{Related Work}

Understanding and detecting suicide risk in online text has been an active research area in recent years. Prior studies have explored traditional machine learning techniques, transformer-based models, and clinically validated scales for mental health assessment. This section discusses (1) suicide risk assessment using NLP, (2) applications of large language models in mental health, and (3) the use of the Columbia-Suicide Severity Rating Scale (C-SSRS) in computational psychiatry.

\subsection{Suicide Risk Assessment in NLP}
Natural Language Processing (NLP) has been widely used to detect signs of suicidal ideation in online forums, social media, and clinical narratives. Early studies employed lexicon-based sentiment analysis to identify suicidal expressions in text \cite{burnap2015machine}. Later, machine learning techniques such as Support Vector Machines (SVMs) and Random Forest classifiers improved suicide risk prediction by incorporating linguistic and psychological features \cite{ji2018supervised, de2013suicidal}.

With the advent of deep learning, transformer-based architectures such as BERT \cite{devlin2019bert} and RoBERTa \cite{liu2019roberta} have demonstrated superior performance in detecting mental health disorders from text. Recent studies apply fine-tuned BERT models on suicide-related datasets to classify posts into different risk levels \cite{matero2019suicide, haque2021deep}. However, these models often lack explainability, raising ethical concerns regarding their real-world application.

\subsection{Large Language Models for Mental Health}
Large Language Models (LLMs) such as GPT-3 and GPT-4 \cite{brown2020language, xu2024mental}  have shown promising results in mental health applications, including sentiment analysis, chatbot-based counseling, and suicide risk prediction. Studies have explored LLMs for generating synthetic mental health dialogues and assisting in psychological assessments \cite{ghanadian2024socially, sharma2024facilitating}. Despite their potential, challenges such as hallucination, bias, and lack of interpretability remain major concerns in deploying LLMs for sensitive tasks like suicide screening.

\subsection{The Columbia-Suicide Severity Rating Scale (C-SSRS) in Computational Psychiatry and Risk Assessment}

\begin{figure}[ht]
    \centering
    \includegraphics[width=0.52\textwidth]{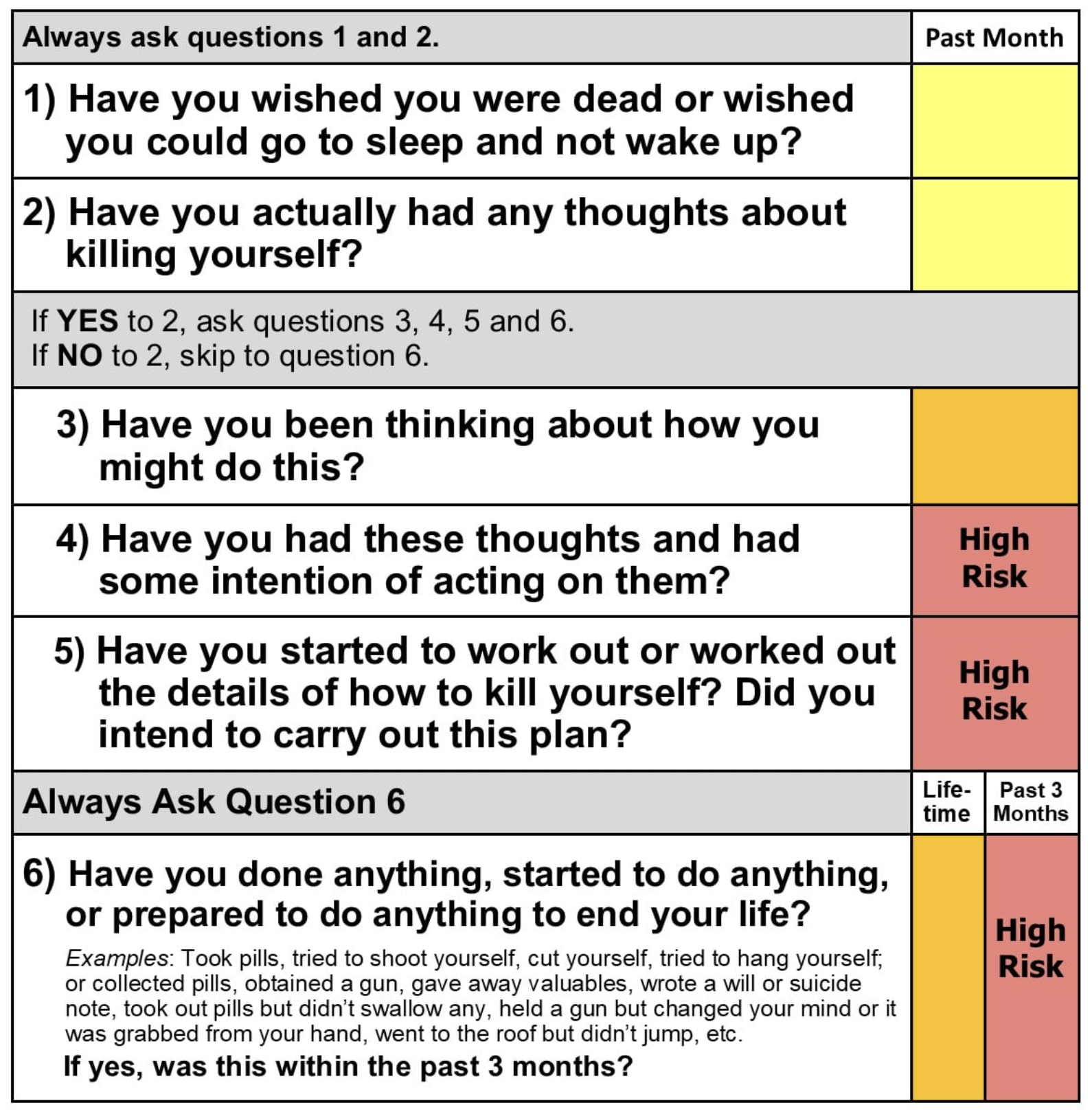}
    \caption{C-SSRS: Suicide Risk Screening Tool}
    \label{fig:cssrs} 
\end{figure}

The Columbia-Suicide Severity Rating Scale (C-SSRS) is a clinically validated tool widely used for suicide risk assessment \cite{posner2011columbia}. As illustrated in Figure \ref{fig:cssrs}, it classifies risk based on the severity of suicidal ideation, intent, and history of previous attempts. Multiple studies have explored mapping social media text to C-SSRS labels due to its wide applicability \cite{gaur2019knowledge} \cite{ji2020suicidal} \cite{patil2025cognitive}. Recent efforts integrate C-SSRS with NLP techniques to develop automated risk assessment systems, but the accuracy and reliability of such approaches remain underexplored.

The C-SSRS consists of a structured set of questions designed to assess suicidal ideation and behaviors. The core questions aim to determine the presence, severity, and progression of suicidal thoughts and actions. Figure \ref{fig:cssrs} provides a representation of the C-SSRS.

By leveraging these structured questions, NLP models can be trained to detect suicide risk from textual data, enhancing computational psychiatry applications. However, challenges persist in ensuring that these automated methods accurately interpret and classify responses.

\section{Dataset}
We collected user-generated posts from the r/SuicideWatch subreddit using the Reddit API \cite{RedditAPI}. These posts reflect individuals’ expressions of suicidal ideation, emotional distress, and personal challenges. Initially, over 3,000 posts were gathered from December 2024. From this set, approximately 1,200 posts were selected for analysis based on specific criteria, which are discussed in a later section. Prior research has established the utility and reliability of Reddit data for mental health studies \cite{de2016discovering, coppersmith2018natural, patil2025cognitive}. In this section, we present insights from our exploratory data analysis and outline the procedures used for data annotation and labeling.

\subsection{Exploratory Data Analysis:}

\begin{figure}[htbp]
    \centering
    \subfloat[Word Cloud Severity 0]{\includegraphics[width=0.48\linewidth]{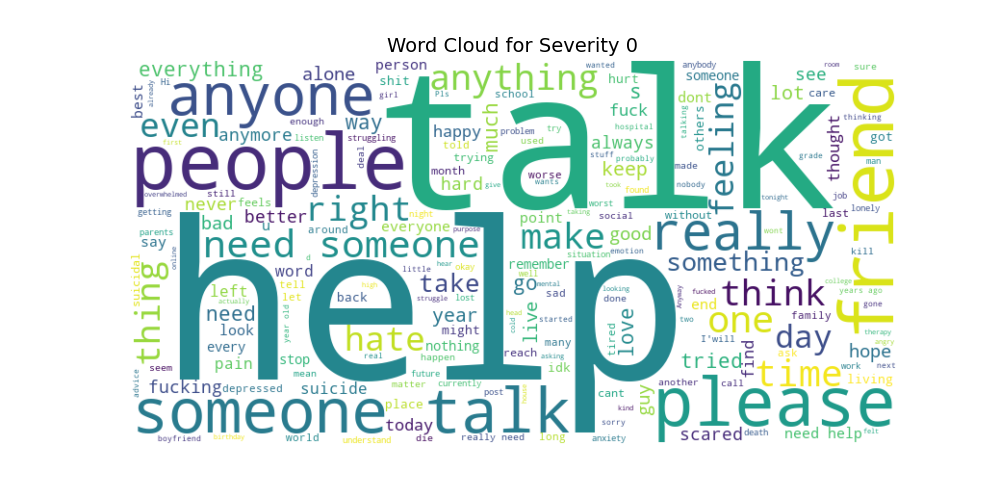}}
    \subfloat[Word Cloud Severity 1]{\includegraphics[width=0.48\linewidth]{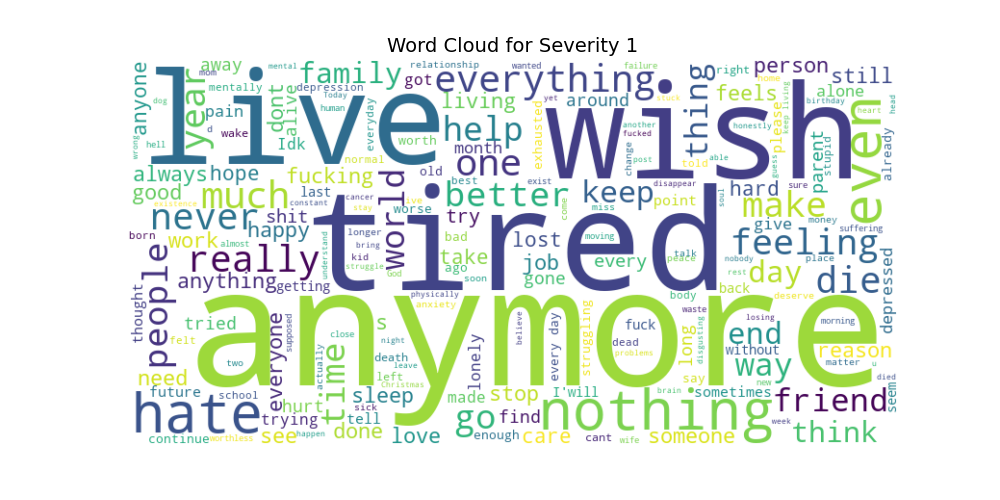}} \\
    \subfloat[Word Cloud Severity 2]{\includegraphics[width=0.48\linewidth]{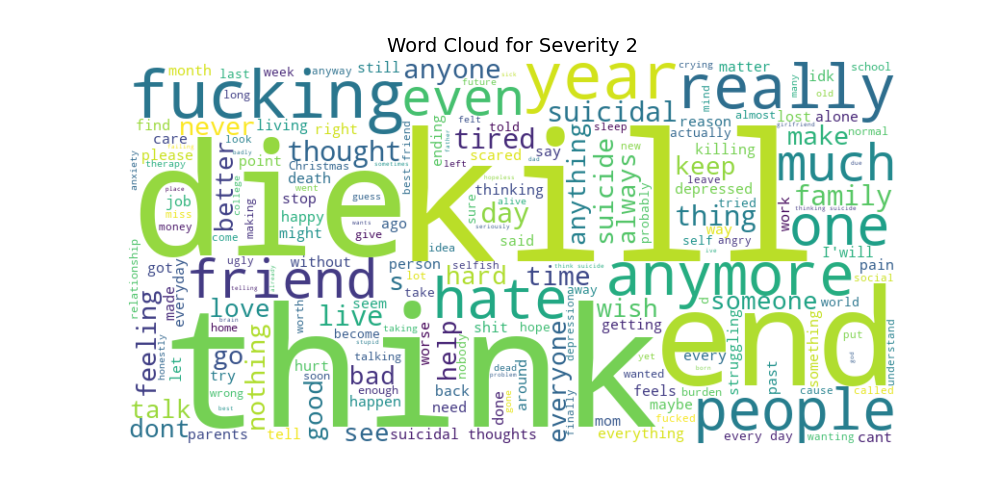}}
    \subfloat[Word Cloud Severity 3]{\includegraphics[width=0.48\linewidth]{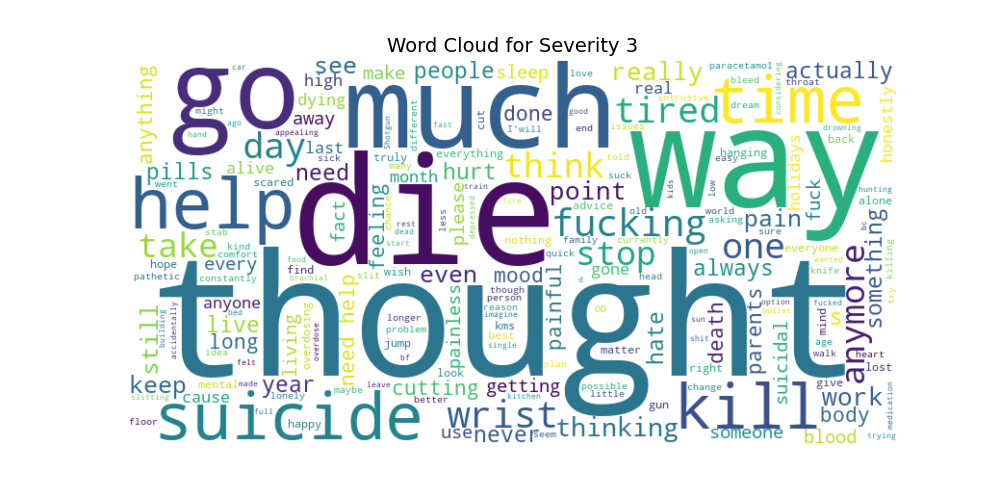}}

    \subfloat[Word Cloud Severity 4]{\includegraphics[width=0.48\linewidth]{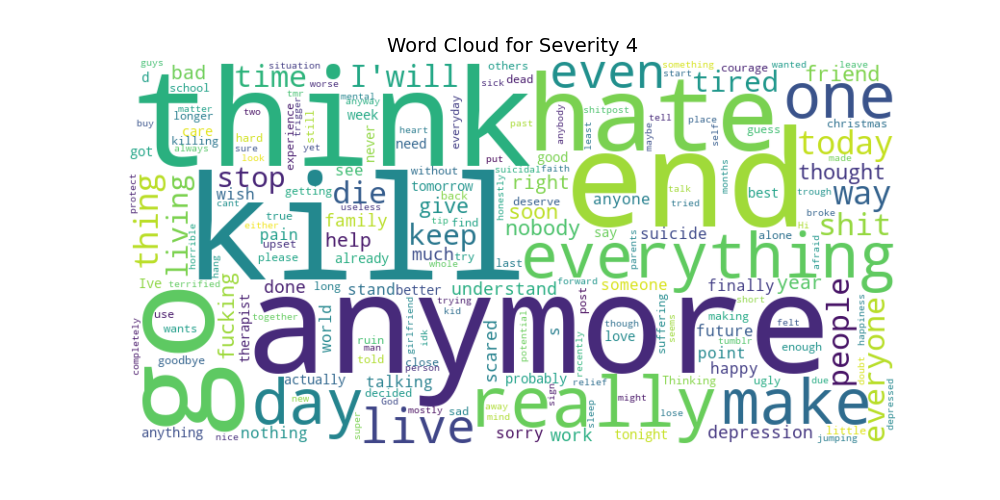}}
    \subfloat[Word Cloud Severity 5]{\includegraphics[width=0.48\linewidth]{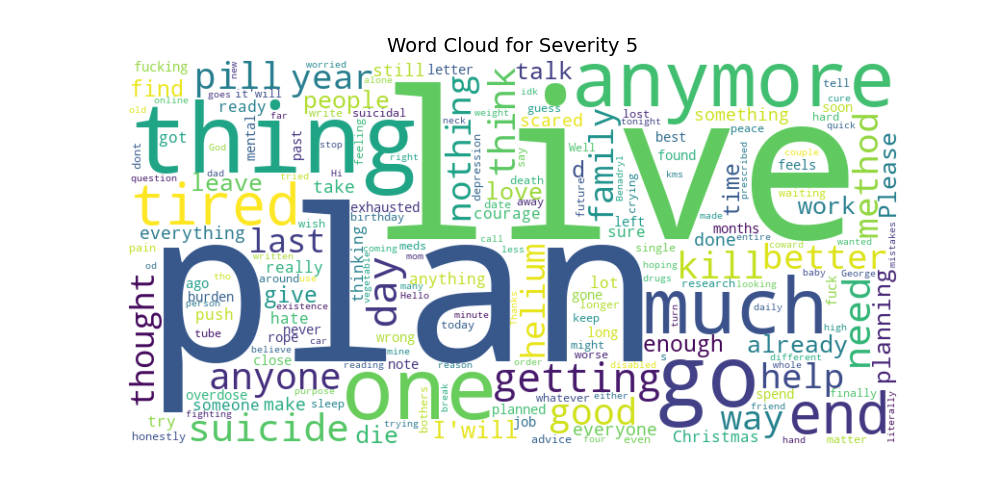}} \\
    \subfloat[Word Cloud Severity 6]{\includegraphics[width=0.48\linewidth]{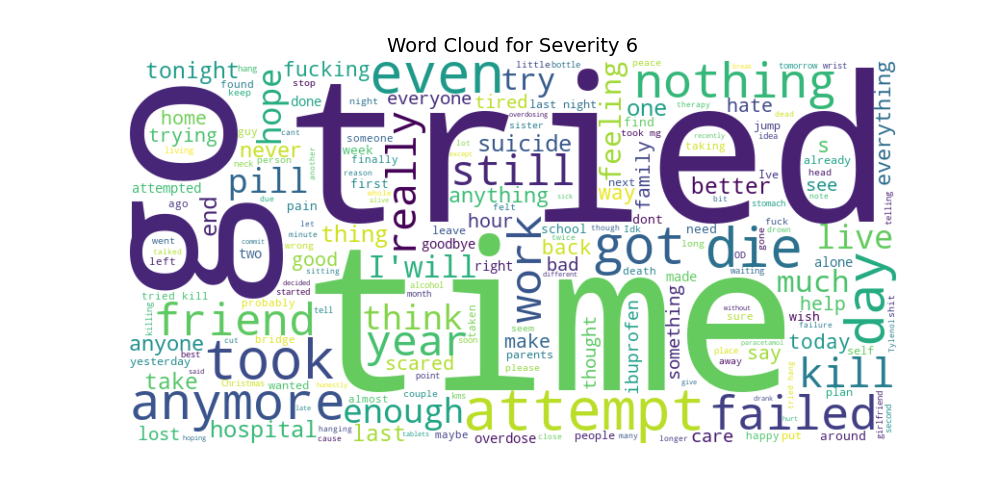}}
    \caption{Word cloud visualizations (Severity 0-6).}
    \label{fig:wordclouds_2}
\end{figure}

\begin{itemize}

\item \textbf{Word Count Distribution by Severity:}
The word count of posts varies significantly across severity levels. The box plot (Figure~\ref{fig:box_word_count}) provides insights into the median, interquartile range (IQR), and potential outliers. Posts with higher severity scores (CSSR-S $\geq 4$) tend to exhibit greater variability in word count, with some containing substantially more words than lower-severity posts. However, lower-severity posts (CSSR-S $\leq 2$) demonstrate a more consistent word count distribution with fewer extreme values.

\begin{figure}[htbp]
\centerline{\includegraphics[width=0.45\textwidth]{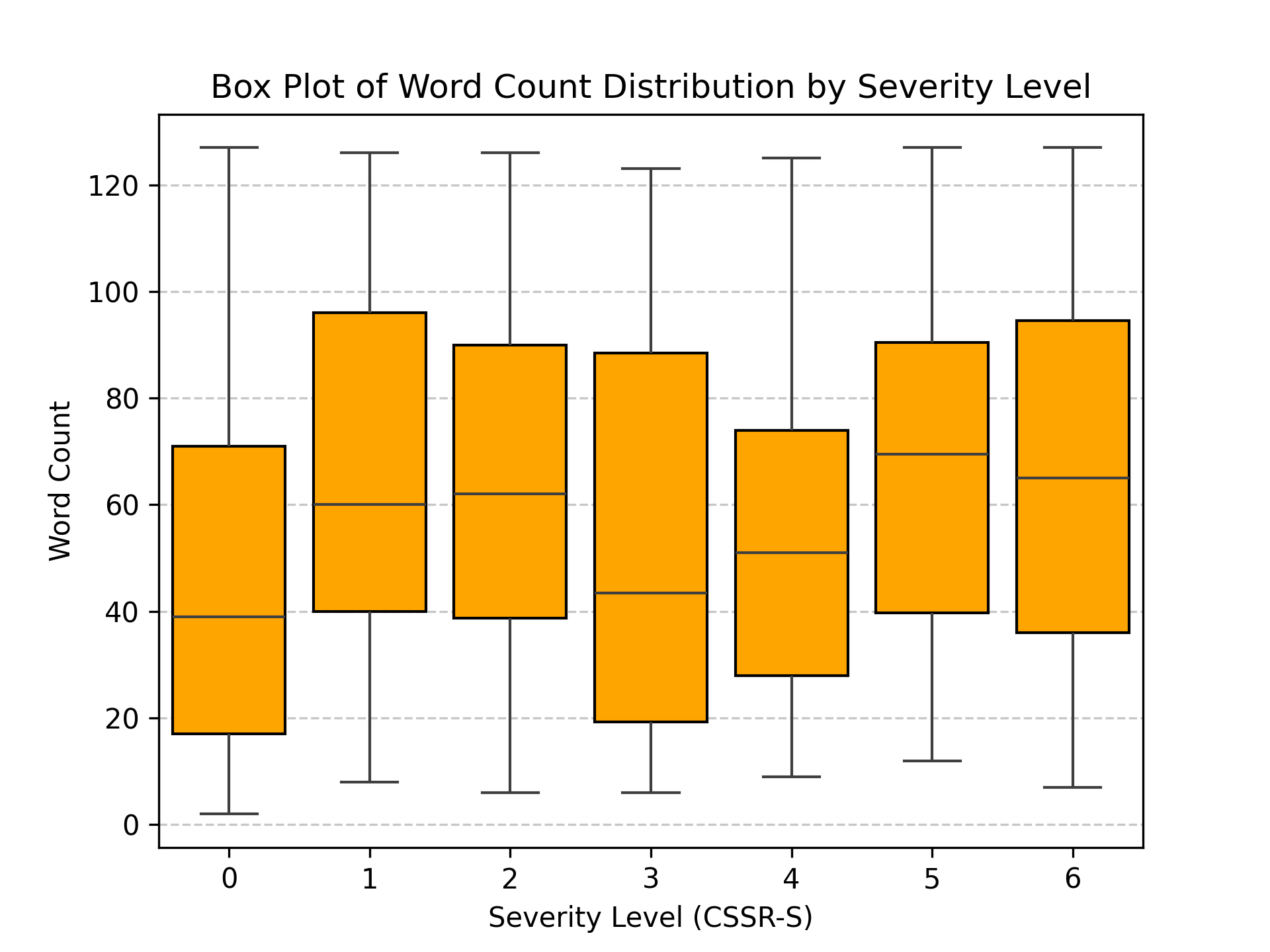}}
\caption{Word Count Distribution By Severity Level}
\label{fig:box_word_count}
\end{figure}

\item \textbf{Daily Posts Frequency:}
The daily posting frequency of the dataset fluctuates over time, as illustrated in Figure~\ref{fig:daily_posts}. Peaks in activity suggest potential external influences, such as seasonal effects or significant events triggering increased engagement on Reddit’s \texttt{SuicideWatch} subreddit. Notably, there are periods with a substantial number of posts, warranting further investigation into temporal patterns.

\begin{figure}[htbp]
\centerline{\includegraphics[width=0.5\textwidth]{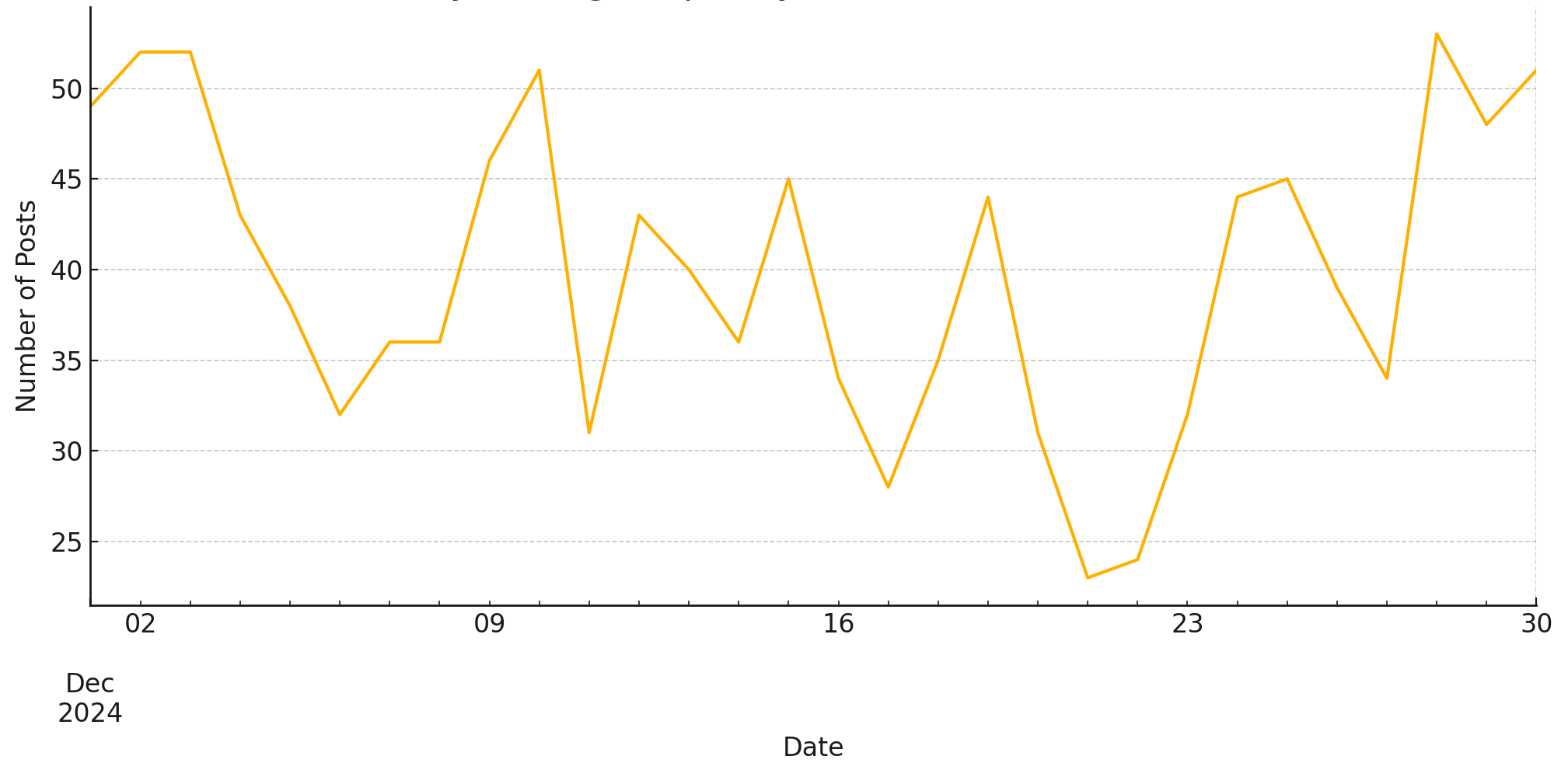}}
\caption{Post Frequency Over Time}
\label{fig:daily_posts}
\end{figure}

\item \textbf{Distribution of Posts per Severity Score:}
The distribution of CSSR-S severity scores is depicted in Figure~\ref{fig:severity_distribution}. The dataset exhibits an imbalance, with certain severity levels being more prevalent. This imbalance could affect the performance of predictive models, necessitating techniques such as data augmentation, weighted loss functions, or resampling methods to ensure model robustness.

\item \textbf{Word Clouds Across Severity Levels:} The word clouds show a progression in language from seeking support (talk, help, friend) at Severity 0 to increasing hopelessness (lonely, tired, suicidal) at Severity 1-2. By Severity 3-4, explicit suicidal ideation emerges (die, wrist, stop, kill), and at Severity 5-6, crisis language dominates with words indicating plans and attempts (plan, attempt, pill, overdose). This trend highlights the shift from emotional distress to imminent risk, useful for suicide risk detection. Figure~\ref{fig:wordclouds_2} represents this linguistic progression, illustrating how distress escalates from general emotional struggles to explicit suicidal intent.

\begin{figure}[htbp]
\centerline{\includegraphics[width=0.5\textwidth]{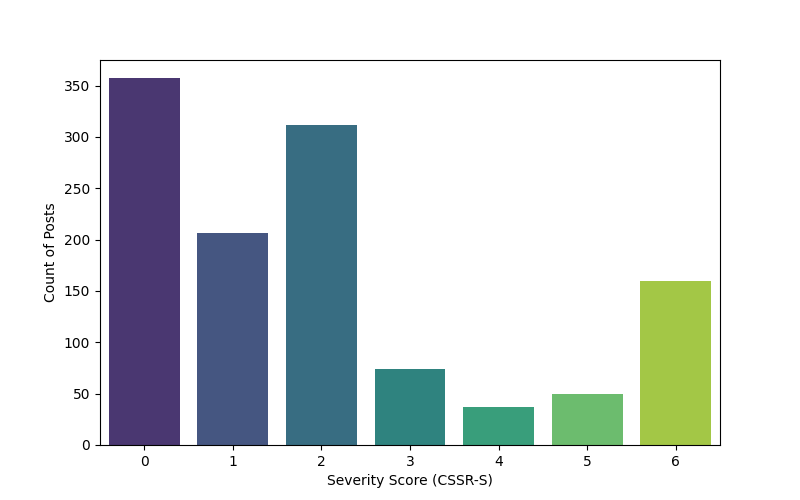}}
\caption{Distribution of Severity Scores}
\label{fig:severity_distribution}
\end{figure}

\end{itemize}

\subsection{Annotation and Labeling}
\textbf{Human Annotation:} A team comprising trained psychologists and two trained C-SSRS assessors manually annotated a subset of 1,200 social media posts. Following the official C-SSRS guidelines \cite{posner2011columbia}, annotators evaluated each post for the severity of suicidal ideation and intent.

The annotation process focused on identifying explicit or implicit expressions of suicidal thoughts, intentions, or behaviors. Posts were examined for specific indicators, including expressions of hopelessness, detailed plans for self-harm, or references to means of suicide. When such content was present, posts were labeled on a scale from 0 to 6, corresponding to the severity levels defined by the C-SSRS.

To ensure consistency, a majority voting mechanism was employed to determine the final label for each post. The annotation process demonstrated strong inter-annotator agreement, with a Cohen’s kappa coefficient of \(\kappa = 0.82\), indicating substantial reliability.

\section{Methodology}
This section details the methodology employed for assessing suicide risk in Reddit’s r/SuicideWatch posts. Our approach comprises four key components: (1) experimental setup, (2) large language model (LLM) systems, and (3) evaluation strategy. We adhered to established protocols from prior CSSR-S studies to ensure reproducibility and rigor~\cite{xu2024mental, Koehn2010}. Posts were preprocessed to remove personal and identifiable information, then annotated with severity labels to serve as ground truth. We evaluated five advanced LLMs—ChatGPT, Google Gemini, Claude, Mistral, and LLaMA—by submitting 1,200 anonymized posts to each system via API using standardized prompts and configurations. This setup enabled a fair, controlled comparison of model performance in the context of suicide risk classification.

\subsection{Experimental Setup}
We followed a systematic process aligned with prior C-SSRS evaluation studies~\cite{xu2024mental, patil2025cognitive, Koehn2010} to ensure reproducibility and reliability:

\begin{enumerate}
    \item \textbf{Preprocessing:} To ensure user anonymity, we removed personally identifiable information (PII), such as names and locations, during preprocessing. The original dataset comprised 3,000 Reddit posts, some of which exceeded practical length constraints. To mitigate the computational demands of large language model (LLM) inference, we limited the dataset to posts with a combined title and body text (selftext) of fewer than 128 words. We also excluded users who posted more than once during the observation period to avoid user-level confounds. While prior studies \cite{gaur2019knowledge} have incorporated temporal or user-level context (e.g., prior posts from the same or related groups) to better differentiate between ideation and intent—a frequent source of misclassification—we found that such additional content often introduced irrelevant noise, occasionally even confounding human annotators. For evaluation, each post was manually annotated with a severity score, which served as the ground truth for computing performance metrics.
    
    \item \textbf{Prompt Design:} 
    To elicit structured and consistent reasoning outputs from the LLM, we designed a multi-part prompt aligned with the Columbia-Suicide Severity Rating Scale (C-SSRS). The prompt instructs the model to evaluate each post step-by-step and assign a severity score on a 0--6 scale, where 0 indicates irrelevance to suicidal ideation and 1--6 correspond to increasing severity levels based on specific C-SSRS questions. The full prompt includes:
    \begin{itemize}
        \item An instruction to assess severity on a defined scale.
        \item A list of six C-SSRS-based criteria, each associated with a severity score from 1 to 6.
        \item A request for structured output in JSON format, with reasoning provided for each question or marked as N/A.
    \end{itemize}
    This design ensures alignment with clinical assessment standards and facilitates both interpretability and quantitative evaluation of model responses. The generated JSON outputs were subsequently parsed for metric computation.

    \item \textbf{Reasoning Strategy:} We evaluated model performance exclusively using chain-of-thought reasoning. Alternative prompting strategies, such as few-shot or instruction tuning variations, were not explored in this study, as they are comprehensively analyzed in a separate work~\cite{patil2025cognitive}.
\end{enumerate}

These components together form the foundation of our LLM-based severity scoring pipeline, as detailed in the subsequent evaluation section.

\subsection{Large Language Models}
\label{sec:translation-systems}

We evaluated five Large Language Reasoning Models, selected based on demonstrated reasoning performance and availability across proprietary and open-weight platforms. These models represent a diverse set of state-of-the-art systems from both commercial and open-source domains, and have shown strong performance in natural language understanding, classification, and inference:

\begin{enumerate}
    \item \textbf{ChatGPT o3-mini:} o3-mini is a lightweight language model developed by OpenAI, optimized for efficiency while retaining strong performance on reasoning tasks. Despite its smaller size compared to models like GPT-4, o3-mini demonstrates robust capabilities in tasks involving logical inference, structured problem-solving, and language understanding.

   \item \textbf{Gemini 1.5 Pro:} Gemini 1.5 Pro is a state-of-the-art multimodal large language model developed by Google DeepMind, capable of handling extended context lengths and complex reasoning tasks. Gemini 1.5 Pro shows strong reasoning abilities, making it well-suited for tasks requiring memory over large inputs and nuanced understanding across modalities \cite{google2024gemini}.
    
    \item \textbf{Claude 3 Sonnet:} Claude 3 Sonnet is a mid-sized model in the Claude 3 family developed by Anthropic. It exhibits strong capabilities in structured reasoning, summarization, and dialogue comprehension, making it well-suited for research and practical applications where efficient yet accurate language understanding is required. Claude 3 Sonnet also reflects Anthropic’s emphasis on model alignment and safety \cite{anthropic2024claude35}.

    \item \textbf{Mistral (Pixtral) Large:} Pixtral Large is a vision-language model developed by Mistral AI. It integrates visual and textual reasoning, enabling it to handle tasks involving image understanding, captioning, and multimodal question answering. Pixtral Large maintains Mistral’s emphasis on efficiency and open accessibility while offering competitive performance in cross-modal reasoning scenarios \cite{mistral2024pixtral}.

    \item \textbf{LLaMA:} LLaMA (Large Language Model Meta AI) is a collection of foundation models developed by Meta. LLaMA 2 and the upcoming LLaMA 3 aim to provide state-of-the-art performance in an open-weight format, encouraging research and application in both academic and industrial settings \cite{touvron2023llama}.
\end{enumerate}

We submitted each of the 1,200 Reddit posts to five different systems via their respective APIs. To ensure fairness in evaluation, all large language models (LLMs) received the same input format, including consistent prompt structures and context settings, and were used with their default configurations. These measures helped mitigate potential biases and ensured a rigorous evaluation of LLM performance in classifying Reddit posts. Additionally, we trained a Support Vector Machine (SVM) text classification model as a baseline for comparison. We selected SVM specifically because it performs well on small to medium-sized datasets and offers a strong traditional benchmark without introducing additional architectural complexity. Other traditional models were not explored in order to maintain the study’s focus on evaluating Large Language Models.

\subsection{Evaluation Metrics}
\label{sec:eval_metrics}

Following best practices in MT research, we employed five automatic metrics to capture different facets of translation quality:

\begin{itemize}
    \item \textbf{Accuracy}\cite{manning2008introduction}: A fundamental metric used to evaluate classification models, measuring the proportion of correctly predicted instances over the total number of instances.
    
    \vspace{6pt}
    \noindent
    \textit{Formulation.}  
    Given a classification task with predictions $\hat{y}$ and ground-truth labels $y$, accuracy is defined in terms of true positives (TP), true negatives (TN), false positives (FP), and false negatives (FN):
    \[
      \mathrm{Accuracy} = \frac{\text{TP} + \text{TN}}{\text{TP} + \text{TN} + \text{FP} + \text{FN}}.
    \]
    This metric provides an overall measure of correctness but may be insufficient for imbalanced datasets. 
    
    For multi-class classification with $N$ samples and $K$ classes, accuracy extends to:
    \[
      \mathrm{Accuracy} = \frac{1}{N} \sum_{i=1}^{N} \mathbb{1}(\hat{y}_i = y_i),
    \]
    where $\mathbb{1}(\cdot)$ is the indicator function that returns 1 if the prediction matches the true label, otherwise 0.

    \item \textbf{Precision, Recall, and F1-score}\cite{manning2008introduction}: These metrics provide deeper insight into classification performance, especially for imbalanced datasets.

    \begin{itemize}
        \item \textbf{Precision (per language)}: Measures how many predicted labels for a given language are actually correct.
        \[
        \mathrm{Precision} = \frac{\text{TP}}{\text{TP} + \text{FP}}.
        \]
        
        \item \textbf{Recall (per language)}: Measures how many actual labels for a given language were correctly predicted.
        \[
        \mathrm{Recall} = \frac{\text{TP}}{\text{TP} + \text{FN}}.
        \]

        \item \textbf{F1-score}: The harmonic mean of precision and recall, balancing both metrics.
        \[
        \mathrm{F1} = 2 \times \frac{\mathrm{Precision} \times \mathrm{Recall}}{\mathrm{Precision} + \mathrm{Recall}}.
        \]

    \end{itemize}

    \item \textbf{Mean Squared Error (MSE)}: Penalizes larger errors more than smaller ones.
    \begin{equation}
        MSE = \frac{1}{n} \sum (y_{\text{true}} - y_{\text{pred}})^2
    \end{equation}

    \item \textbf{Mean Absolute Error (MAE)}: Measures the average absolute difference between predicted and true severity.
    \begin{equation}
        MAE = \frac{1}{n} \sum |y_{\text{true}} - y_{\text{pred}}|
    \end{equation}

    \item \textbf{Quadratic Weighted Kappa (QWK)}: Measures agreement between predictions and ground truth, penalizing bigger mistakes more heavily. Useful for ordinal data.

    \begin{equation}
        \kappa = 1 - \frac{\sum_{i,j} O_{ij} W_{ij}}{\sum_{i,j} E_{ij} W_{ij}}
    \end{equation}

    \item \textbf{Confusion Matrix}: A structured table that helps visualize how often each language is correctly classified or misclassified as another. The confusion matrix for $K$ classes is defined as:
    \[
    C_{i,j} = \text{Instances of class } i \text{ classified as class } j.
    \]
    The diagonal elements represent correct predictions, while off-diagonal elements indicate misclassifications.

    \item \textbf{Spearman’s Rank Correlation:} Checks if model predictions preserve the ranking order of severity levels. 
    \[
    \rho = 1 - \frac{6 \sum d_i^2}{n(n^2 - 1)}
    \]


\end{itemize}

\section{Results}

This section presents the evaluation results of our zero shot LLM-based suicide risk classifier using the Columbia-Suicide Severity Rating Scale (C-SSRS). The evaluation includes standard classification metrics (accuracy, precision, recall, F1-score) as well as regression-based error measures (MAE, MSE) and ordinal agreement metrics (Quadratic Weighted Kappa [QWK] and Spearman’s Rank Correlation [SRC]).

\subsection{Overall Performance}

Table~\ref{tab:performance} presents the performance metrics of the selected Large Language Models (LLMs)—Claude, GPT, LLaMA, Mistral, and Gemini—on the C-SSRS severity classification task. 

\begin{table}
    \centering
    \caption{Comparison of Model Performance Metrics (Sorted by F1-Score)}
    \label{tab:performance}
    \begin{tabular}{lrrrrrr}
        \toprule
        {} & Claude & GPT & Mistral & LLaMA & Gemini & SVM \\
        \midrule
        Accuracy                       & \textbf{0.7331} & 0.7278 & 0.7132 & 0.6560 & 0.5958 & 0.5583 \\
        Precision \textsuperscript{*} & \textbf{0.7880} & 0.7818 & 0.7478 & 0.7477 & 0.6901 & 0.5437 \\
        Recall \textsuperscript{*}    & \textbf{0.7331} & 0.7278 & 0.7132 & 0.6560 & 0.5958 & 0.5583 \\
        F1-Score \textsuperscript{*}  & \textbf{0.7505} & 0.7384 & 0.7208 & 0.6839 & 0.6160 & 0.5385 \\
        MAE                           & 0.4502 & 0.4884 & \textbf{0.4398} & 0.6049 & 0.6285 & 0.8958 \\
        MSE                           & 1.0427 & 1.1481 & \textbf{1.0067} & 1.3633 & 1.3565 & 2.6792 \\
        QWK                           & 0.8758 & 0.8692 & \textbf{0.8767} & 0.8444 & 0.8322 & 0.6502 \\
        SRC                           & 0.8809 & \textbf{0.8843} & 0.8640 & 0.8766 & 0.8308 & 0.6137 \\
        \bottomrule
    \end{tabular}

    \vspace{0.5em}
    \begin{minipage}{\linewidth}
        \small
        \textit{MAE:} Mean Absolute Error; 
        \textit{MSE:} Mean Squared Error; 
        \textit{QWK:} Quadratic Weighted Kappa; 
        \textit{SRC:} Spearman’s Rank Correlation. 
        \textsuperscript{*}Weighted average across classes.
    \end{minipage}
\end{table}

Claude achieved the highest overall performance across most evaluation metrics, including an accuracy of \textbf{73.31\%}, a precision of \textbf{78.80\%}, a recall of \textbf{73.31\%}, and an F1-score of \textbf{75.05\%}. It also maintained a high Quadratic Weighted Kappa (QWK) of \textbf{0.8758}, indicating strong agreement with human annotations on ordinal labels.

Mistral, while ranking third in F1-score (\textbf{72.08\%}), demonstrated the lowest Mean Absolute Error (\textbf{0.4398}) and Mean Squared Error (\textbf{1.0067}), as well as the highest QWK (\textbf{0.8767}), suggesting excellent alignment with ground truth labels in ordinal prediction tasks. This highlights its effectiveness in minimizing prediction error, even if slightly behind Claude in classification metrics.

GPT closely followed Claude, with an F1-score of \textbf{73.84\%}, accuracy of \textbf{72.78\%}, and the highest Spearman’s Rank Correlation (SRC) of \textbf{0.8843}, suggesting strong performance in rank-order predictions.

In contrast, LLaMA and Gemini performed moderately, with LLaMA reaching an F1-score of \textbf{68.39\%} and Gemini trailing at \textbf{61.60\%}. SVM showed the weakest performance overall, underscoring the advantage of LLMs in this task.

These results indicate that advanced LLMs—particularly Claude and GPT—are well-suited for C-SSRS-based suicide risk classification. Mistral also offers competitive performance, especially in reducing ordinal prediction errors.

\subsection{Confusion Matrix Analysis}

\begin{figure*}[t]
    \centering
    \begin{subfigure}[b]{0.32\textwidth}
        \includegraphics[width=\textwidth]{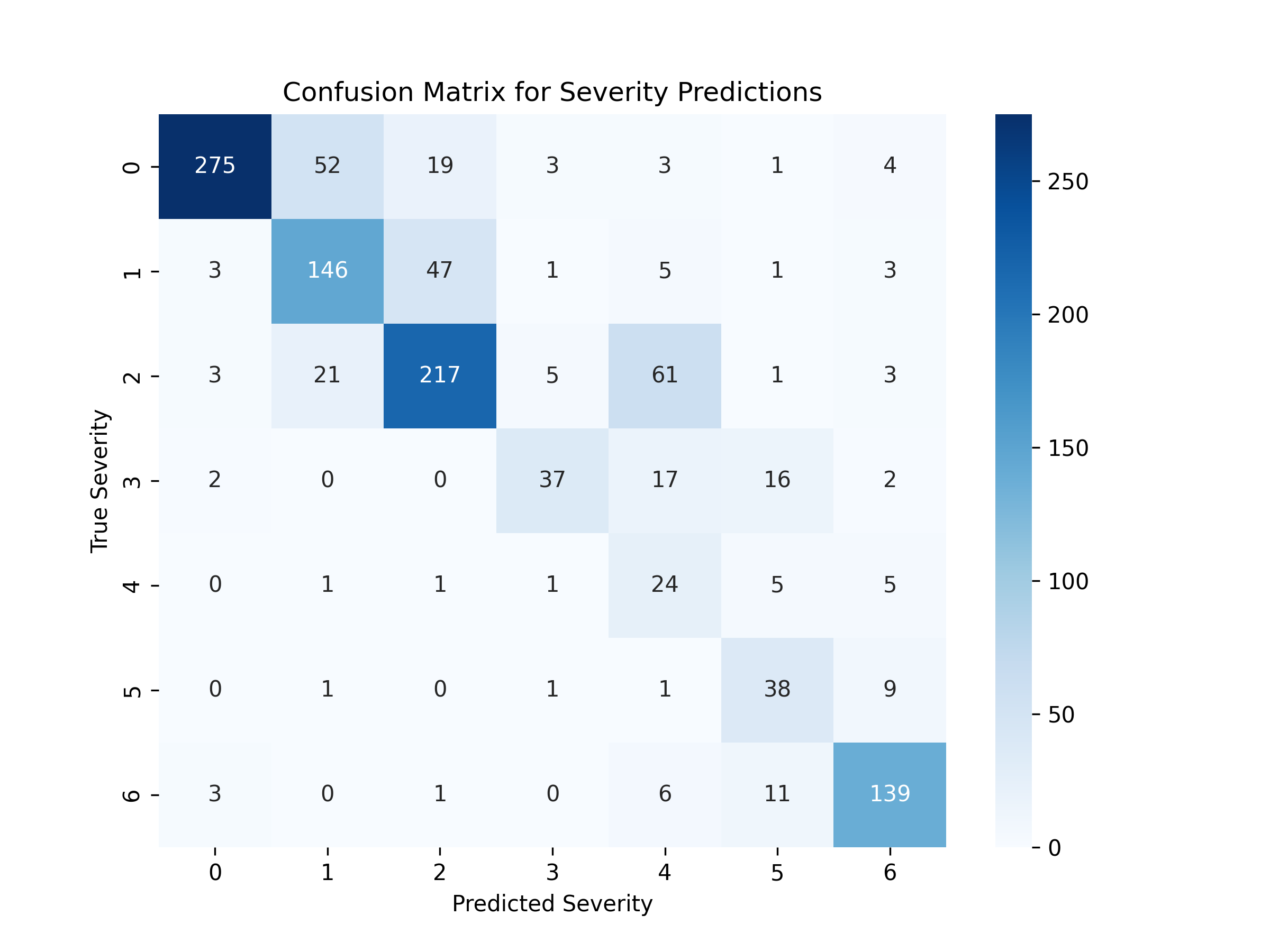}
        \caption{Claude}
        \label{fig:claude_cm}
    \end{subfigure}
    \hfill
    \begin{subfigure}[b]{0.32\textwidth}
        \includegraphics[width=\textwidth]{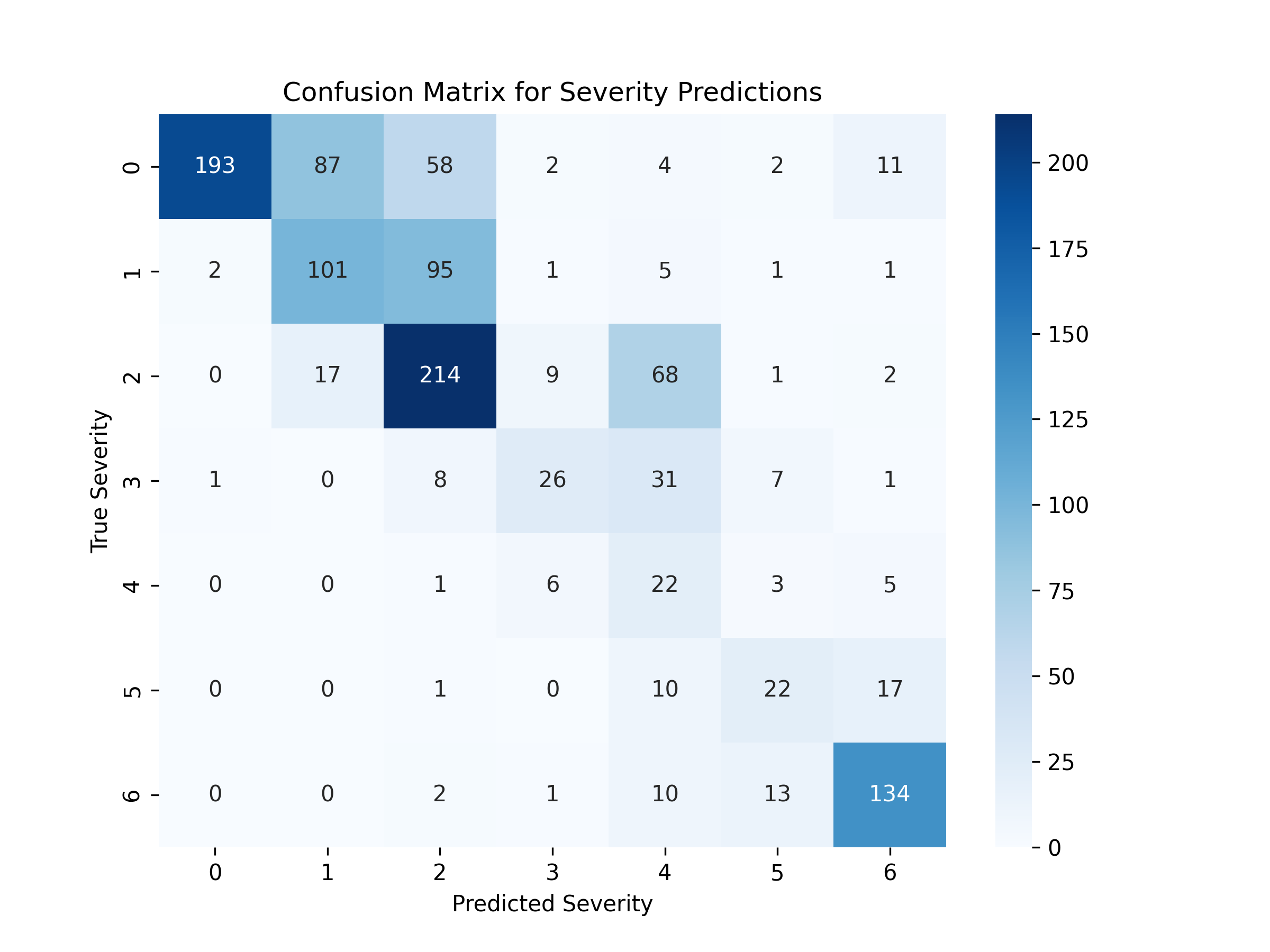}
        \caption{Gemini}
        \label{fig:gemini_cm}
    \end{subfigure}
    \hfill
    \begin{subfigure}[b]{0.32\textwidth}
        \includegraphics[width=\textwidth]{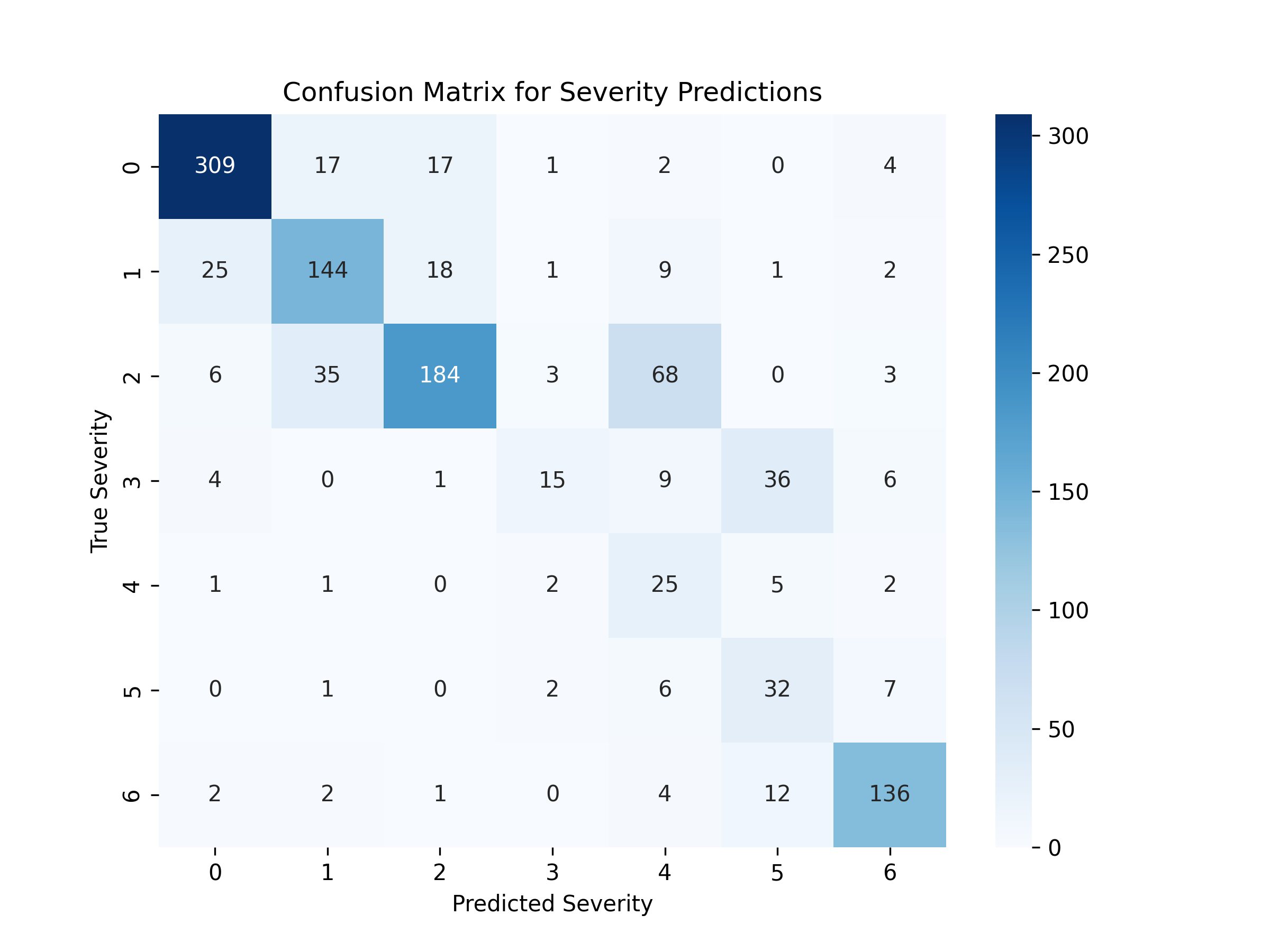}
        \caption{GPT}
        \label{fig:gpt_cm}
    \end{subfigure}
    
    \vspace{0.2cm} 
    \begin{subfigure}[b]{0.32\textwidth}
        \includegraphics[width=\textwidth]{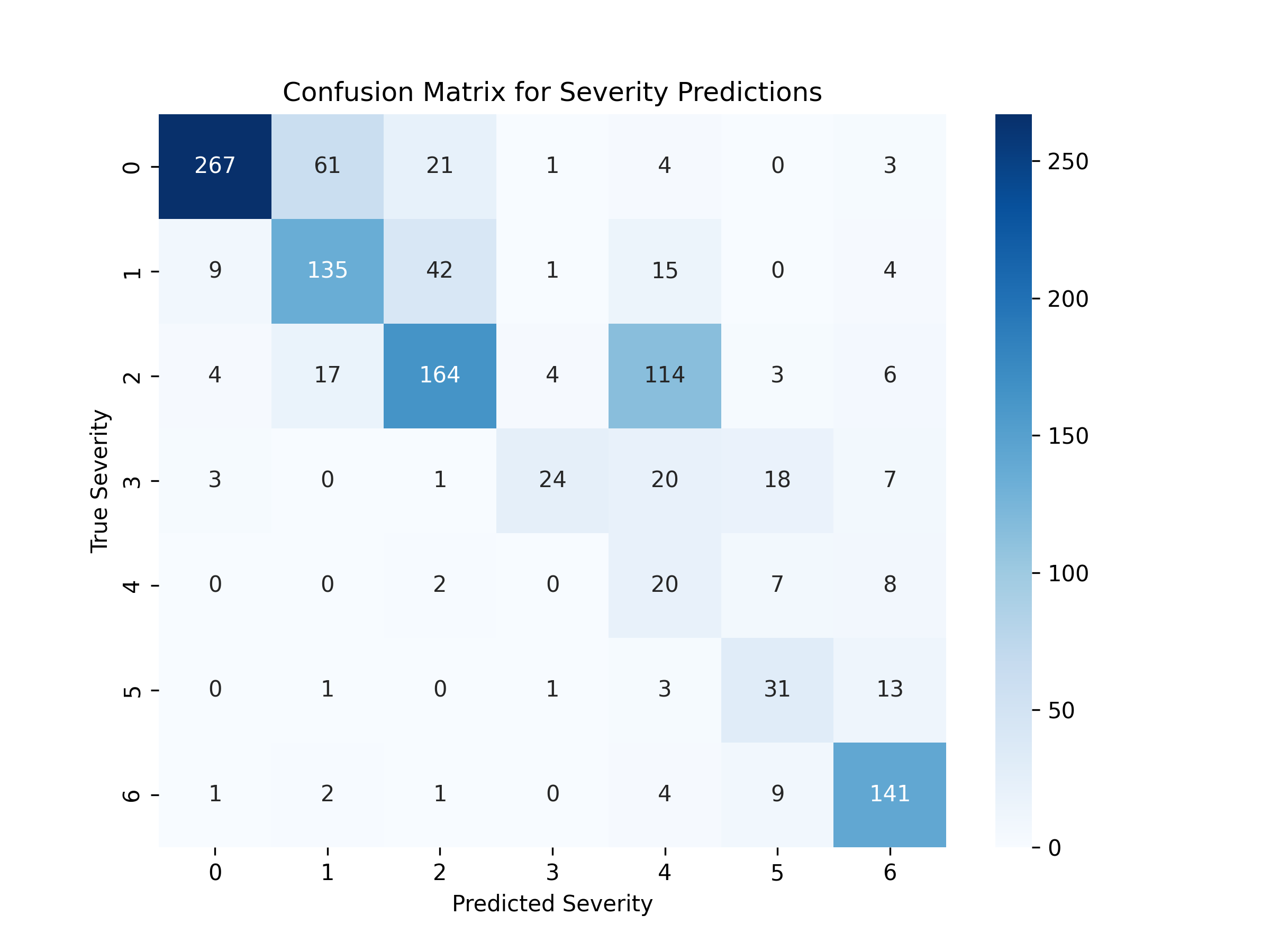}
        \caption{Llama}
        \label{fig:llama_cm}
    \end{subfigure}
    \hfill
    \begin{subfigure}[b]{0.32\textwidth}
        \includegraphics[width=\textwidth]{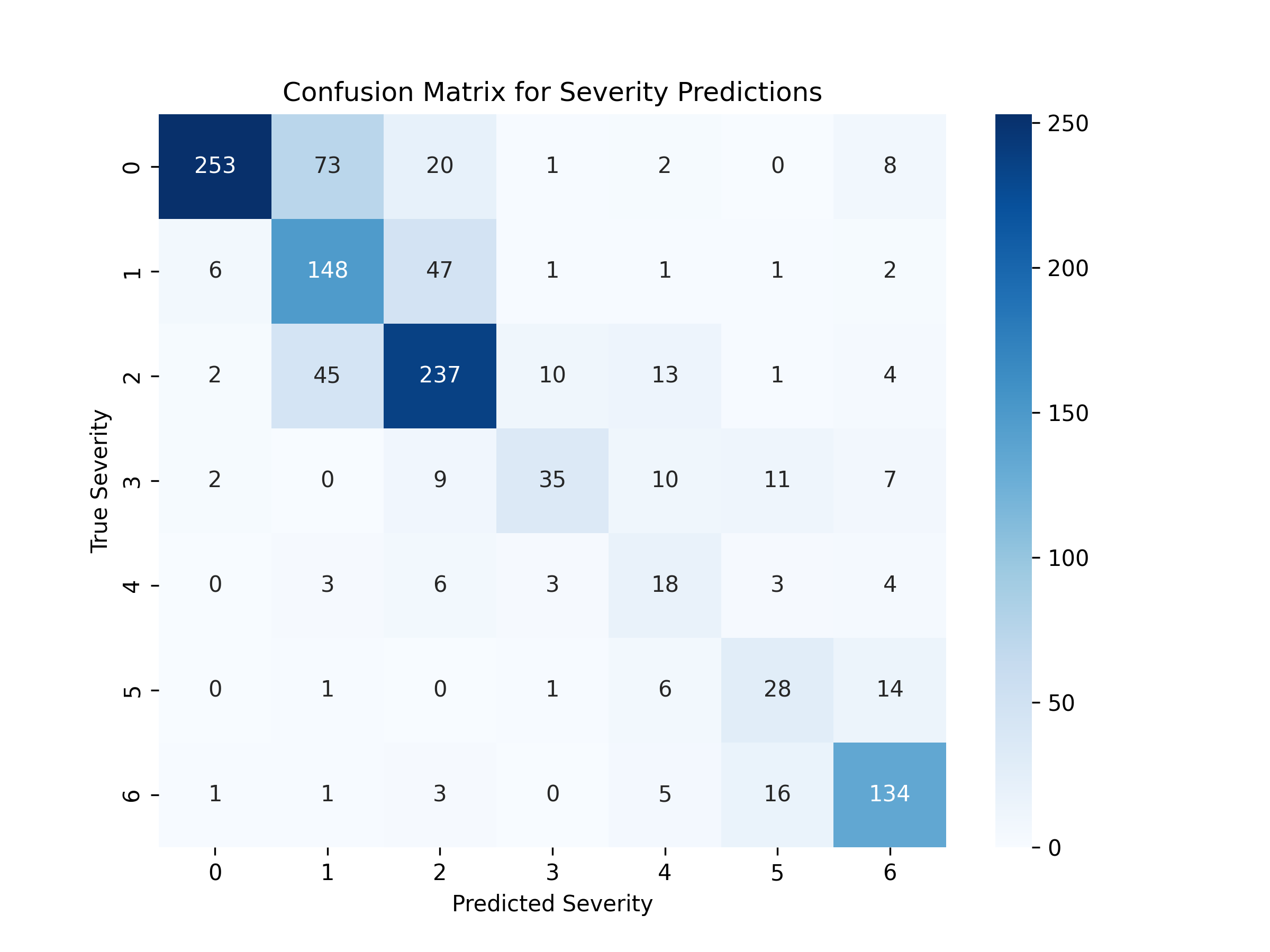}
        \caption{Mistral}
        \label{fig:mistral_cm}
    \end{subfigure}
    \hfill
    \begin{subfigure}[b]{0.32\textwidth}
        \includegraphics[width=\textwidth]{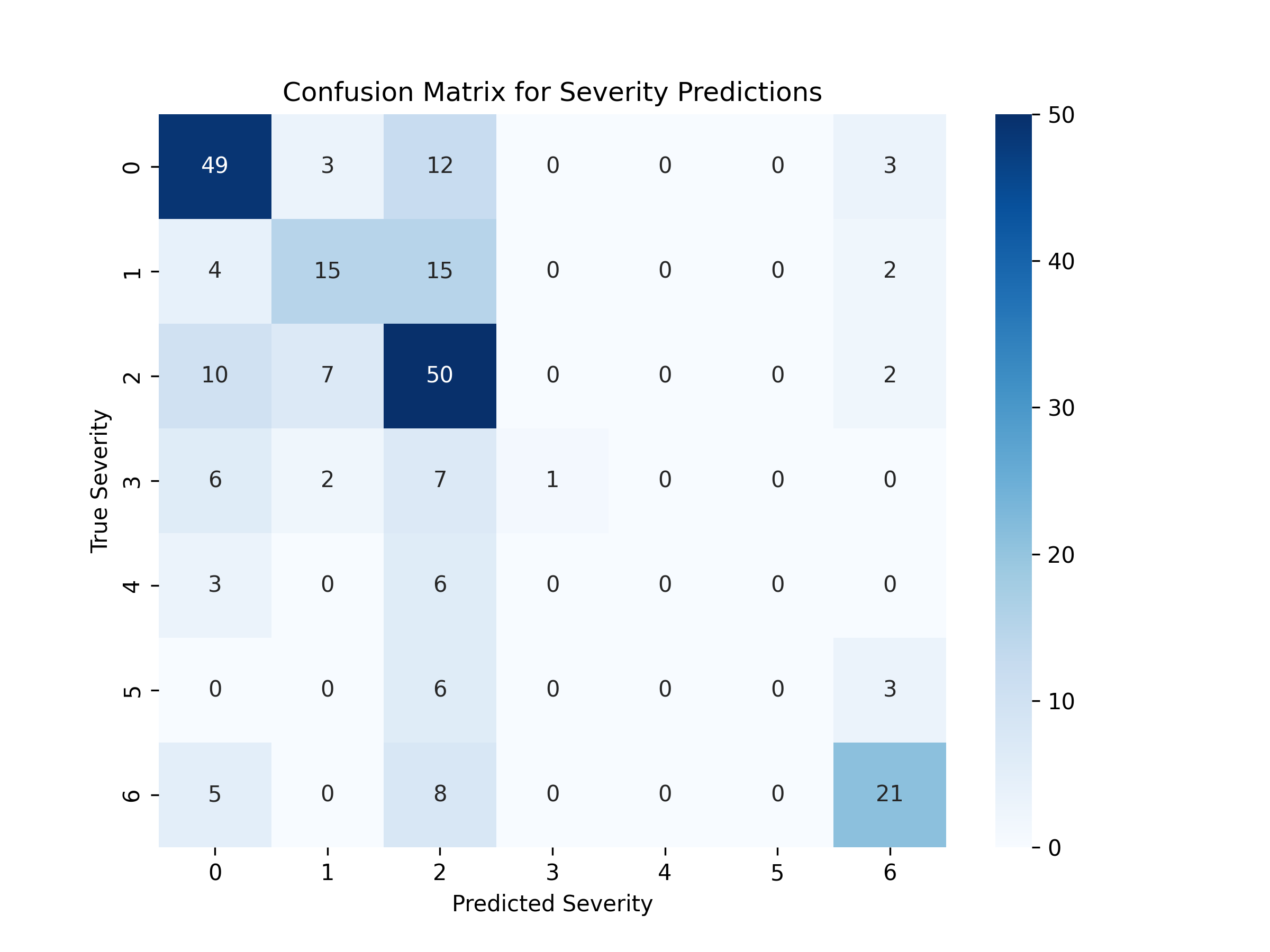}
        \caption{SVM}
        \label{fig:svm_cm}
    \end{subfigure}
    
    \caption{Confusion matrices for language identification across seven machine translation models. Darker diagonal entries indicate correct classifications.}
    \label{fig:confusion_matrices}
\end{figure*}

Figure~\ref{fig:confusion_matrices} presents the confusion matrices for each model’s predictions across the 0–6 C-SSRS severity levels. A consistent trend across all models is that misclassifications predominantly occur between adjacent severity levels, suggesting that while exact class predictions may be incorrect, the ordinal relationships are generally preserved. This behavior implies that the models are sensitive to the underlying severity gradient, even when precise classification is missed.

Claude and GPT demonstrated the strongest alignment with ground truth, reflected in their dominant diagonals and minimal off-diagonal dispersion. Claude showed particularly strong performance in correctly identifying Level 0 and Level 2 posts, with most misclassifications concentrated between Level 1 and Level 2. GPT's matrix was similar but exhibited slightly more spread across Levels 0–2, which corresponds to its slightly lower recall compared to Claude.

Mistral, while not leading in F1-score, performed well in Levels 2 through 4, indicating accurate mid-range severity prediction. However, it struggled to distinguish clearly between Levels 3 and 5. This aligns with its low Mean Absolute Error (MAE), indicating that while exact classes may not always match, predicted values remain numerically close to the actual severity levels.

LLaMA showed greater confusion among early severity levels, especially between Level 0 and Levels 1 and 2, which may explain its weaker classification metrics. Errors appeared less symmetric, suggesting some inconsistency in learning ordinal boundaries.

Gemini had the highest degree of dispersion, particularly across Levels 0–2 and Level 4, indicating a lack of clarity in differentiating between adjacent and distant severity levels. This aligns with its relatively low overall accuracy, QWK, and Spearman correlation.

SVM exhibited the most pronounced confusion, with substantial misclassification even among non-adjacent severity levels. Its matrix lacked a strong diagonal structure, reinforcing the limitations of traditional machine learning approaches on this ordinal, nuanced task.

These findings underscore that while all models benefit from ordinal awareness, the degree of misclassification and its alignment with severity proximity varies by model architecture. LLMs like Claude and GPT appear better aligned with the structure of C-SSRS, while others show varying degrees of degradation in finer-grained predictions.

\subsection{Error Analysis}

Across all models, most classification errors involved mislabeling posts as adjacent C-SSRS severity levels, reinforcing the challenge of capturing subtle distinctions in language indicative of suicide risk progression.

\textbf{Levels 3 vs. 4 (Active Suicidal Ideation with vs. without Intent):} Misclassifications were particularly common between these two levels, especially in GPT and Claude. This likely stems from the nuanced language used to convey intent, which may not be explicitly stated in user posts. For instance, expressions of hopelessness or desperation can be interpreted in multiple ways depending on phrasing and context, making intent detection especially difficult.

\textbf{Levels 5 vs. 6 (Plan vs. Attempt):} Several models—including Gemini and LLaMA—frequently confused Level 5 (Intent with Plan) and Level 6 (Actual Attempt). These errors often occurred in cases where posts described suicidal thoughts and planning but omitted or obscured any mention of a prior attempt. Such distinctions may require deeper inference or contextual understanding that current models struggle to extract reliably.

\textbf{Level 0 vs. Levels 1 or 2:} Mistral and Gemini showed elevated false positives for early-stage ideation in posts labeled as Level 0 (No Suicidal Ideation). This indicates that these models may be overly sensitive to emotionally charged language, even when explicit suicidal ideation is not present. Such sensitivity, while cautious, may lead to overclassification in clinical contexts.

These observations highlight the difficulty of reliably distinguishing between neighboring severity levels in C-SSRS classification and underscore the importance of incorporating richer contextual understanding and improved intent recognition in future model development.

\subsection{Ethical and Practical Implications}

Despite high accuracy and alignment with human annotations, deploying LLM-based classifiers in mental health contexts presents serious ethical and practical challenges. These models must serve solely as assistive tools—augmenting, not replacing, clinical judgment or moderator review.

Particular concern arises from false negatives, where high-risk posts (e.g., Levels 5–6) may be underestimated, delaying crucial intervention. Conversely, false positives can cause unnecessary distress or escalations.

To mitigate such risks, we advocate for human-in-the-loop oversight and tiered intervention systems that combine automated triage with expert validation. Transparent disclosure of model limitations, decision auditability, and informed user consent should be integral to any deployment pipeline.

\subsection{Summary of Findings}

This study evaluated the performance of several state-of-the-art LLMs in zero-shot suicide risk assessment on r/SuicideWatch posts using the Columbia-Suicide Severity Rating Scale (C-SSRS). Key findings include:

\begin{itemize}
    \item \textbf{Performance:} Claude and GPT achieved the best overall accuracy and F1 scores, closely aligning with human annotations. Mistral showed strong ordinal calibration with the lowest MAE and MSE, despite slightly lower classification metrics.

    \item \textbf{Error Patterns:} Misclassifications often occurred between adjacent levels (e.g., 1–2, 3–4, 5–6), reflecting the continuum of suicidal ideation rather than random errors.

    \item \textbf{Input Sensitivity:} Short or context-poor posts were more prone to misclassification. Posts requiring inference about intent or action (e.g., plan vs. attempt) were particularly challenging.

    \item \textbf{Ethical Use:} Given the high stakes, LLM outputs should be treated as decision-support tools, not substitutes for human judgment.
\end{itemize}

These results highlight both the promise and the limits of LLMs in sensitive classification tasks.

\section{Discussion and Conclusion}

This study assessed the zero-shot capabilities of Large Language Models (LLMs) to classify suicide risk in Reddit’s r/SuicideWatch posts using the C-SSRS framework. Claude and GPT performed closest to human annotations, while Mistral showed strong ordinal calibration despite lower classification accuracy.

Most errors occurred between adjacent severity levels, reflecting the continuum of suicidal ideation rather than arbitrary misclassification. Short, ambiguous posts and subtle linguistic cues—such as intent versus action—proved especially challenging. These findings highlight both the promise and the nuance of applying LLMs in mental health contexts.

While zero-shot prompting is efficient, it may underperform relative to fine-tuned or few-shot methods \cite{patil2025cognitive}. Further, Reddit-specific language may limit generalizability, and LLMs remain vulnerable to training biases and cultural blind spots—posing risks in high-stakes settings like suicide prevention.

We advocate for LLMs as assistive, not autonomous, tools. Deployed responsibly—with human oversight, auditability, and contextual integration—they can support triage efforts and reduce moderator burden in high-volume platforms.

In sum, LLMs show potential to enhance digital mental health workflows. Future work should prioritize interpretability, contextual adaptation, and cross-platform validation to ensure safe and effective deployment.

\bibliographystyle{IEEEtran}
\bibliography{ref}
\vspace{12pt}

\vspace{12pt}

\newpage
\appendix
\section*{Appendix A: Full LLM Responses for Selected \texttt{r/SuicideWatch} Posts}

This appendix provides the full output of several Large Language Models (LLMs)—Claude, GPT, Gemini, Mistral, and LLaMA—on three posts from Reddit's \texttt{r/SuicideWatch}, using a standardized zero-shot prompt aligned with the Columbia-Suicide Severity Rating Scale (C-SSRS).

\subsection*{Selected Reddit Posts}

\textbf{Post 1 — User: -Arry-}
\begin{quote}
\small
I can’t do this anymore. Last year I tried taking a whole bottle of pills and I didn’t succeed. Please I don’t want to hear it about how much I’m loved or how much things are going to get better. It’s bullshit. Don’t shame me for what I choose to do. It’s my wish, it’s my life and I no longer want to be here. I can’t take it anymore. Does anyone know what pills will actually do the job? Or maybe how many pill bottles? Please.
\end{quote}

\textbf{Post 2 — User: \#NAME?}
\begin{quote}
\small
I seriously considered committing for a second.... I've never seriously considered killing myself before today and it was only for a second. I need someone who doesn't know my sister to talk to—it's her fault, but I don’t have any friends who don’t like her so I can't talk to them about it.... I'm not actively suicidal right now I just want to move away from my parents and start a new life somewhere else away from all this. I need new friends and maybe even a significant other. I hate being a child and I hate my life.
\end{quote}

\textbf{Post 3 — User: -PatkaLopikju-}
\begin{quote}
\small
I love you guyssss I feel so spiny I'm so happy let's all be happy I love you reddit.
\end{quote}

\subsection*{Prompt Given to All LLMs}

\begin{quote}
\small
Evaluate this post step by step and assign them severity on a scale of 0–6, where 0 means the post is not related to suicide, and 1–6 depending on the metrics below where 1 is for the first question and 6 for the last.

COLUMBIA-SUICIDE SEVERITY RATING SCALE metrics:

1) Have you wished you were dead or wished you could go to sleep and not wake up? Severity: 1

2) Have you actually had any thoughts about killing yourself? Severity: 2

3) Have you been thinking about how you might do this? Severity: 3

4) Have you had these thoughts and had some intention of acting on them? Severity: 4

5) Have you started to work out or worked out the details of how to kill yourself? Did you intend to carry out this plan? Severity: 5

6) Have you done anything, started to do anything, or prepared to do anything to end your life? Severity: 6

Provide your response in JSON format, using the following template:

\begin{verbatim}
{
  "Q1": Reasoning or "N/A",
  "Q2": Reasoning or "N/A",
  "Q3": Reasoning or "N/A",
  "Q4": Reasoning or "N/A",
  "Q5": Reasoning or "N/A",
  "Q6": Reasoning or "N/A",
  "severity": integer (0–6)
}
\end{verbatim}
\end{quote}

\subsection*{Post 1: LLM Responses}

\paragraph{Claude}
\begin{verbatim}
{
  "Q1": "I no longer want to be here",
  "Q2": "I can't do this anymore",
  "Q3": "Does anyone know what pills will actually do the job?",
  "Q4": "I can't take it anymore",
  "Q5": "Details on pill dosage sought",
  "Q6": "Last year I tried taking a whole bottle of pills",
  "severity": 6
}
\end{verbatim}

\paragraph{GPT}
\begin{verbatim}
{
  "Q1": "Expresses wish to not be alive",
  "Q2": "Ongoing thoughts of ending life",
  "Q3": "Asks for suicide methods",
  "Q4": "Indicates intent",
  "Q5": "Working out plan details",
  "Q6": "Mentions previous attempt",
  "severity": 6
}
\end{verbatim}

\paragraph{Mistral}
\begin{verbatim}
{
  "Q1": "I can’t do this anymore",
  "Q2": "No longer want to be here",
  "Q3": "Don't shame me for my choice",
  "Q4": "I can't take it anymore",
  "Q5": "Pill method exploration",
  "Q6": "Previous failed attempt",
  "severity": 6
}
\end{verbatim}

\paragraph{Gemini}
\begin{verbatim}
{
  "Q1": "States not wanting to live",
  "Q2": "Previous suicide attempt",
  "Q3": "Looking for methods",
  "Q4": "Implied intent",
  "Q5": "Still exploratory",
  "Q6": "No current prep",
  "severity": 4
}
\end{verbatim}

\paragraph{LLaMA}
\begin{verbatim}
{
  "Q1": "Wishes not to be alive",
  "Q2": "Has had suicidal thoughts",
  "Q3": "Considering methods",
  "Q4": "Some intention to act",
  "Q5": "Working out details",
  "Q6": "No current prep evidence",
  "severity": 5
}
\end{verbatim}

\subsection*{Post 2: LLM Responses}

\paragraph{Claude}
\begin{verbatim}
{
  "Q1": "I hate my life",
  "Q2": "Brief consideration of suicide",
  "Q3": "N/A",
  "Q4": "N/A",
  "Q5": "N/A",
  "Q6": "N/A",
  "severity": 2
}
\end{verbatim}

\paragraph{GPT}
\begin{verbatim}
{
  "Q1": "N/A",
  "Q2": "Mentions suicidal thought",
  "Q3": "N/A",
  "Q4": "Not actively suicidal",
  "Q5": "N/A",
  "Q6": "N/A",
  "severity": 2
}
\end{verbatim}

\paragraph{Mistral}
\begin{verbatim}
{
  "Q1": "I hate my life",
  "Q2": "Brief consideration of suicide",
  "Q3": "N/A",
  "Q4": "N/A",
  "Q5": "N/A",
  "Q6": "N/A",
  "severity": 2
}
\end{verbatim}

\paragraph{Gemini}
\begin{verbatim}
{
  "Q1": "N/A",
  "Q2": "Fleeting suicidal thought",
  "Q3": "N/A",
  "Q4": "No current intent",
  "Q5": "N/A",
  "Q6": "N/A",
  "severity": 2
}
\end{verbatim}

\paragraph{LLaMA}
\begin{verbatim}
{
  "Q1": "Implied dissatisfaction",
  "Q2": "Brief suicidal ideation",
  "Q3": "N/A",
  "Q4": "No active intention",
  "Q5": "N/A",
  "Q6": "N/A",
  "severity": 2
}
\end{verbatim}

\subsection*{Post 3: LLM Responses}

\paragraph{All Models (Claude, GPT, Mistral, Gemini, LLaMA)}
\begin{verbatim}
{
  "Q1": "N/A",
  "Q2": "N/A",
  "Q3": "N/A",
  "Q4": "N/A",
  "Q5": "N/A",
  "Q6": "N/A",
  "severity": 0
}
\end{verbatim}

\end{document}